\documentclass[letterpaper, 10 pt, conference]{ieeeconf}  

\IEEEoverridecommandlockouts                              

\usepackage{graphicx} 
\usepackage{subcaption}
\usepackage{multirow}

\title{\LARGE \bf
Performance Evaluation of Low-Cost Machine Vision Cameras for Image-Based  Grasp Verification
}

\author{Deebul Nair$^{1}$,  Amirhossein Pakdaman$^{1}$ and Paul G. Pl{\"o}ger$^{1}$
\thanks{$^{1}${Deebul Nair, Amirhossein Pakdaman and Prof. Dr. Paul G. Pl{\"o}ger are with the  Dept.  of  Computer  Science,  Bonn-Rhein-Sieg  University  of  Applied Sciences, Germany. \small deebul.nair@h-brs.de,  amirhossein.pakdaman@smail.inf.h-brs.de, paul.ploeger@h-brs.de}}%
}
\usepackage{todonotes}
\begin{document}

\maketitle
\thispagestyle{empty}
\pagestyle{empty}

\begin{abstract}
Grasp verification is advantageous for autonomous manipulation robots as they provide the feedback required for higher level planning components about successful task completion. However, a major obstacle in doing grasp verification is sensor selection.  In this paper, we propose a vision based grasp verification system using machine vision cameras, with the verification problem formulated as an image classification task. Machine vision cameras consist of a camera and a processing unit capable of on-board deep learning inference. The inference in these low-power hardware are done near the data source, reducing the robot’s dependence on a centralized server, leading to reduced latency, and improved reliability. Machine vision cameras provide the deep learning inference capabilities using different neural accelerators. Although, it is not clear from the documentation of these cameras what is the effect of these neural accelerators on performance metrics such as latency and throughput. 
To systematically benchmark these machine vision cameras, we propose a parameterized model generator that generates end to end models of Convolutional Neural Networks(CNN). Using these generated models we benchmark latency and throughput of two machine vision cameras, JeVois A33 and Sipeed Maix Bit.
Our experiments demonstrate that the selected machine vision camera and the deep learning models can robustly verify grasp with 97\% per frame accuracy. 

\end{abstract}

\section{INTRODUCTION}

\begin{figure}[t]
      \centering
      \includegraphics[width=0.9\linewidth]{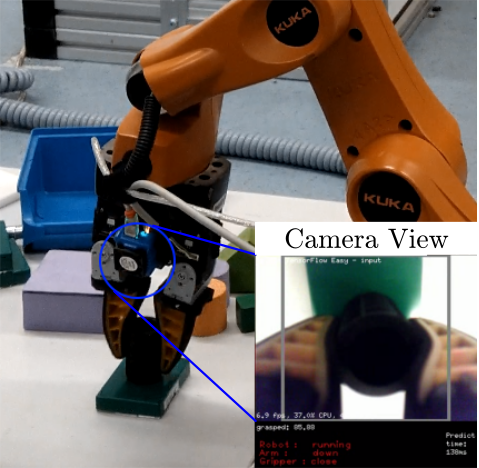}
      \caption{youBot gripper with machine vision camera JeVois A33 along with the camera view.} 
      \label{arm_cams}
\end{figure}

Grasp verification is a necessitate for autonomous robots to determine the state of the grasp while performing object manipulation. 
Normally robots need to perform a series of operations that depend on each other. If a task is not done correctly then the robot should re-build the operation sequence based on the occurred failure. So the robot needs to verify its action to update its knowledge about the current state. Grasp verification in robots, generally, are done using sensor's in the gripper but this becomes challenging in the new flexible grippers. For example, in the KUKA youBot(see fig \ref{arm_cams}) we have added a parallel adaptive gripper fingers by Festo. The adaptive nature of the gripper finger makes it difficult for placing traditional sensors and to robustly determine the state of the grasp. In this paper, we propose a \textit{machine vision camera} sensor based grasp verification that works by capturing images of the gripper and verifying if the grasp is successful using deep learning inference.

A machine vision camera consists of an image sensor, a processor and a neural processing unit, this makes it possible to perform edge computing on the captured image. Performing inference near the data source reduces the load to transfer and process images on the centralized server. This helps in reducing the latency in processing the data and improves reliability of the overall system. Recently, deep neural networks had great improvements in solving image classification problems. This was achieved by the advent of convolutional neural networks (CNNs). Successful CNN approaches are able to solve image classification problems with an accuracy near to human \cite{canziani2017evaluation}. However, CNNs need considerable resources of computation and memory.
The idea of executing a deep learning algorithm into an embedded device has been discussed widely and solutions such as compression are provided \cite{alizadeh2018empirical, howard2017mobilenets}. In this paper, we formulate the grasp verification problem as deep learning based image classification task. However, the two main challenges with performing deep learning inference in machine vision cameras are (i) lack of literature on the performance of these cameras and (ii) establishing a deep learning architecture that is capable to fulfill the visual classification task with satisfying accuracy reckoning the limitations of the embedded device. 

In this paper, we benchmark the performance of machine vision cameras by creating a parameterized model generator which generates CNN models of varying parameters, executing them and recording performance metrics.
Based on the hardware-software limitation of the machine vision cameras, the benchmarking results and the grasp verification task we select corresponding deep learning models. Finally all the grasp verification system is integrated which includes dataset collection, training and deployment of machine vision camera compatible deep learning models. To summarize, the main contributions of the paper are the \textbf{performance evaluation} of low-cost machine vision cameras and integration of machine vision camera with a real world robot for \textbf{grasp verification} tasks. We hope that the performance evaluation will drive the robotic community to apply machine vision cameras in other robotics applications.

\section{RELATED WORK}
Several researchers have studied autonomous grasping and vision based grasping. They all have pointed the importance of grasp verification in terms of the demand for autonomous capabilities and for adapting manipulation in dynamic environments. The method presented in \cite{stansfield1991robotic} is a general-purpose robotic grasping system that is designed to work in unstructured environments. Motions of the arm and fingers are automatically generated and validated by extracting a set of features from the environment. Visual information is obtained by scanning the target object from different views. Obtained visual perceptions are processed to generate and execute the grasping task. The visual processing is also used to detect the valid grasp which is done passively.

Machine learning methods and deep learning methods were also explored. Applying deep learning methods for grasp detection is widely used in \cite{caldera2018review}. However, deploying visual learning approaches particularly for grasp verification is not considered.
A grasp system for under-actuated robotic hands is presented in \cite{yao2009analysis}. The method outputs a grasp strategy for each object which is based on the analyses of human knowledge. Valid grasps are detected and used to control the robotic hand. This examination is done by using a well-trained neural network. The attribute parameters of the object are extracted and applied as inputs to the neural network. The result of the network is then compared with the grasp strategy decision.
The approach presented in \cite{seredynski2015grasp} is a grasp planning procedure that extracts a sequence of desired poses from the object as well as expected external forces applied to the object during the task execution. The control system presented in this work contains a grasp verification step that checks the stability of the grasp after the task execution. The approach presented in \cite{kulkarni2019low} uses proximity sensors for grasp detection where the gripper is equipped with flexible fingers. Proximity sensors are used to measure deformation of flexible fingers due to external force and is used to detect a grasp. 

\section{MACHINE VISION CAMERAS}
The machine vision cameras studied in this work are Sipeed Maix Bit, JeVois A33, and OpenMV H7. Hardware aspects define capabilities on computation power, response speed, and communication capabilities. Software aspects define capabilities on which types of neural network layers and activation functions they can accept.  
\begin{figure}[t]
      \centering
      \includegraphics[scale=0.06]{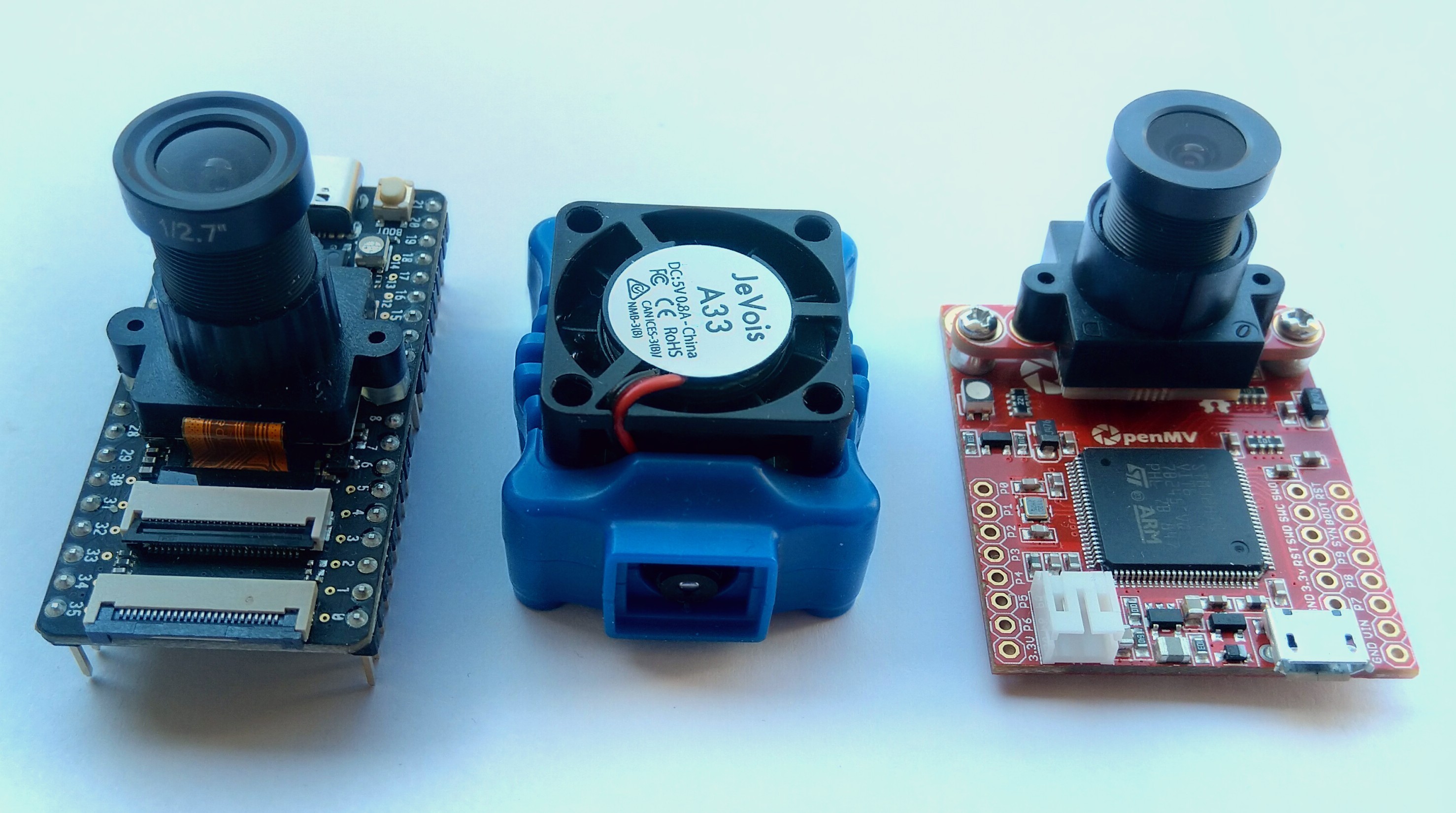}
      \caption{Machine vision cameras. From left to right: Sipeed Maix Bit, JeVois A33, OpenMV H7}
      \label{cams}
\end{figure}

\subsection{JeVois A33}
JeVois runs linux OS that is flashed on the micro SD memory. The firmware contains libraries such as OpenCV 4.0, TensorFlow Lite, Darknet deep neural networks, DLib, etc.
JeVois can run codes written in C++ and Python.
Running the 1.3 GHz CPU requires a large current flow in a smaller footprint which causes overheat issues and consequently, the need for the cooling fan. Thus, this camera consumes more power than others.
The supported file format for CNN models is \textit{tflite} and all layers, activation types of TensorFlow Lite are supported.

\subsection{Sipeed Maix Bit}
Sipeed Maix Bit supports MaixPy programming language which is the MicroPython language that is ported to the K210 processor. Frequently-used standard libraries plus some custom libraries, are available in MaixPy.
It executes a specific format model file called \textit{kmodel}. A neural network compiler called \textit{nncase}\footnote{https://github.com/kendryte/nncase} converts TFLite and caffe models to corresponding kmodel format. The accelerator currently only has support for a restricted group of CNN layers and activation functions also the size of the \textit{kmodel} is limited to 5.9MiB \footnote{https://maixpy.sipeed.com/en/libs/Maix/kpu.html}.

\subsection{OpenMV H7}
OpenMV camera supports MicroPython, a compact implementation of Python 3.4. OpenMV supports only a group of Python functions and class libraries.
This camera accepts only model files with a special file format (\textit{.network}). This type of file can be created by converting CNN models that are created by \textit{Caffe} framework.
The main drawback of this device arises when trying to execute custom deep learning models. After modifying the CNN architecture the conversion procedure fails as its currently not supported. 

\subsection{Comparison}
Even though all these cameras have capability to perform visual computing and execute deep learning models they also have differences such as processor speed, memory, power consumption, and most importantly,the neural accelerators and their capability to do deep learning inference.
Table \ref{table_cameras} displays general specifications of the studied machine vision cameras.
\begin{table}[h]
	\caption{Hardware specifications of the cameras}
	\label{table_cameras}
	\begin{center}
	\begin{tabular}{|l|c|c|c|}
		\hline
		Parameter & JeVois & Sipeed & OpenMV \\
		\hline \hline
		Processor    & ARM Cortex A7  & Kendryte K210 & Arm Cortex-M7 \\
		& 4$\times$1.3 GHz & 400 MHz                & 400 MHz              \\
		& 32 Bit & 64 Bit & 32 Bit  \\ \hline
		Options      & GPU$^1$                & KPU$^2$         & FPU$^3$                 \\ \hline
		RAM          & 256MB              & 8MB                    & 1MB                  \\ \hline
		Comm.        & Serial over USB    & USB to TTL             & USB Streaming        \\ 
		& Micro Serial                    & JTAG, ISP              & SPI, I2C, UART       \\ \hline
		Size [mm]    & 40 x 32 x 21       & 54 x 26 x 13           & 45 x 36 x 30         \\ \hline
		Power        & 5V, $>$800 mA      & 5V, $>$600 mA          & 3.3V, 170 mA         \\ \hline
		Image Size   & $1280\times1024$   & $1632\times1232$       & $640\times480$       \\ \hline
	\end{tabular}
	\end{center}
	\footnotesize{$^1$Dual core Mali-400, $^2$Kendryte Processing Unit (NN Processing Unit), $^3$Floating-point Processing Unit}
\end{table}

\begin{table}[h]
	\caption{Software specifications of the cameras}
	\label{table_cameras_software}
	\begin{center}
	\begin{tabular}{|l|c|c|c|}
		\hline
		Parameter & JeVois & Sipeed & OpenMV \\
		\hline \hline
		Firmware                        & Linux       & MicroPython & Micropython \\ \hline
		Language                        & Python, C++ & MaixPy & MicroPython \\ \hline
		File format                     & .tflite     & .kmodel & .network \\ \hline
        Accept CNNs                    & $\surd$     & $\surd$  & $\surd$  \\ \hline
		Accept custom CNNs             & $\surd$ & $\surd$  & $\times$ \\ \hline
		large model size                & $\surd$ & $\times$ & $\times$ \\
		\hline
	\end{tabular}
	\end{center}
\end{table}

Table \ref{table_cameras_software} compares parameters that are essential for the task of CNN model inference.
Based on the hardware software comparison, JeVois and Sipeed camera were selected for performance analysis. OpenMV camera was dropped because of the software limitation to convert custom deep learning models required for the performance analysis.

\section{PERFORMANCE ANALYSIS}
\begin{figure}[t]
		\includegraphics[width=\linewidth]{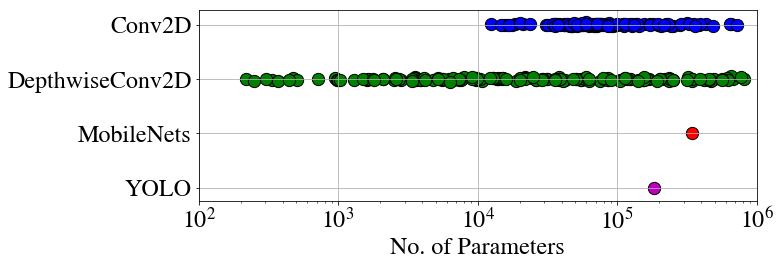}
		\caption{Distribution of number of parameters of the benchmarked models}
		\label{fig_bench_sizes}
	\end{figure}

We benchmark the cameras by executing different custom CNN models and standardized of-the-shelf deep learning architecture. In the first benchmark, we use a CNN model generator which generates a range of CNN models with varying number of parameters while in the second benchmark we use two standardized deep learning models MobileNet\cite{howard2017mobilenets} and YOLO\cite{redmon2017yolo9000}. For each experiment, the models perform inference on-board and two different performance metrics are recorded: \textit{latency} and \textit{throughput}. Fig \ref{fig_bench_sizes} shows the distribution of all individual experiments with respect to the number of parameters.

\textit{Latency} is computed as the 95th percentile of each inference’s timing window, or the time taken for one inference to complete  and \textit{Throughput} is the average inference performance computed  by  the total number of inferences  performed within the sum of all  timing  windows in our case the number of inference performed per second \footnote{https://github.com/eembc/mlmark}. 

\subsection{CNN Model Generator}
CNN model generator, generates a range of CNN models by iterating through specified ranges of hyper-parameters. The resulted models vary in size, number of trainable parameters, and attributes. 
Two groups of CNN models are created for this experiment: one with \textit{Conv2D} layers and the other with \textit{DepthwiseConv2D} layers.
Each model is described by the number of blocks, number of filters (only for Conv2D models), number of inputs (image size), and number of outputs.
Different combinations of these hyper-parameters result in a range of models that are created and used for the experiment. 
Table \ref{table_conv2dparam} and \ref{table_depconv2dparam} shows the range of hyper-parameters used.
\begin{table}[h]
	\caption{Hyper-parameters for Conv2D models}
	\label{table_conv2dparam}
	\begin{center}
	\begin{tabular}{|l|c|c|c|}
		\hline
        Parameter & Start & End & Increment \\
		\hline \hline
        Blocks  & 2  & 6    & 1   \\ \hline
		Filters & 34 & 42   & 4   \\ \hline
		Images  & 16 & 224  & 52  \\ \hline
		Outputs & 2  & 10   & 4   \\
		\hline
	\end{tabular}
	\end{center}
\end{table}
\begin{table}[h]
	\caption{Hyper-parameters for DepthwiseConv2D models}
	\label{table_depconv2dparam}
	\begin{center}
	\begin{tabular}{|l|c|c|c|}
		\hline
        Parameter & Start & End & Increment \\
		\hline \hline
        Blocks  & 1  & 5    & 1   \\ \hline
		Images  & 64 & 224  & 32  \\ \hline
		Outputs & 2  & 82   & 16  \\
		\hline
	\end{tabular}
	\end{center}
\end{table}

To choose these values, constraints of the devices are taken into consideration. Besides, ranges are selected such that resulting models cover a complete set with various file sizes and specifications. The first constraint is about the resulting file size that is imposed by Sipeed camera. The second constraint is that Sipeed cannot accept layer outputs that are lower than $4\times4$ in size. Since the number of outputs decreases as layers are added to a CNN architecture, equation \ref{eq1} defines a relation between hyper-parameters of input size and number of blocks.
\begin{equation}
    \frac{i}{2^{b}} \geq 4 
    \label{eq1}
\end{equation}
Where $i$ is input size (images) and $b$ is the number of blocks. Another constraint is the maximum size of the input image that is again imposed by Sipeed. Based on the experiments, it can only handle models with an input image size of $224\times224$ pixels or less. 
In order to achieve more comparable values, experiments on JeVois camera are done on the two processor frequencies of 1344 MHz and 408 MHz.

\begin{figure}[h!]
	\centering
	\includegraphics[width=\linewidth]{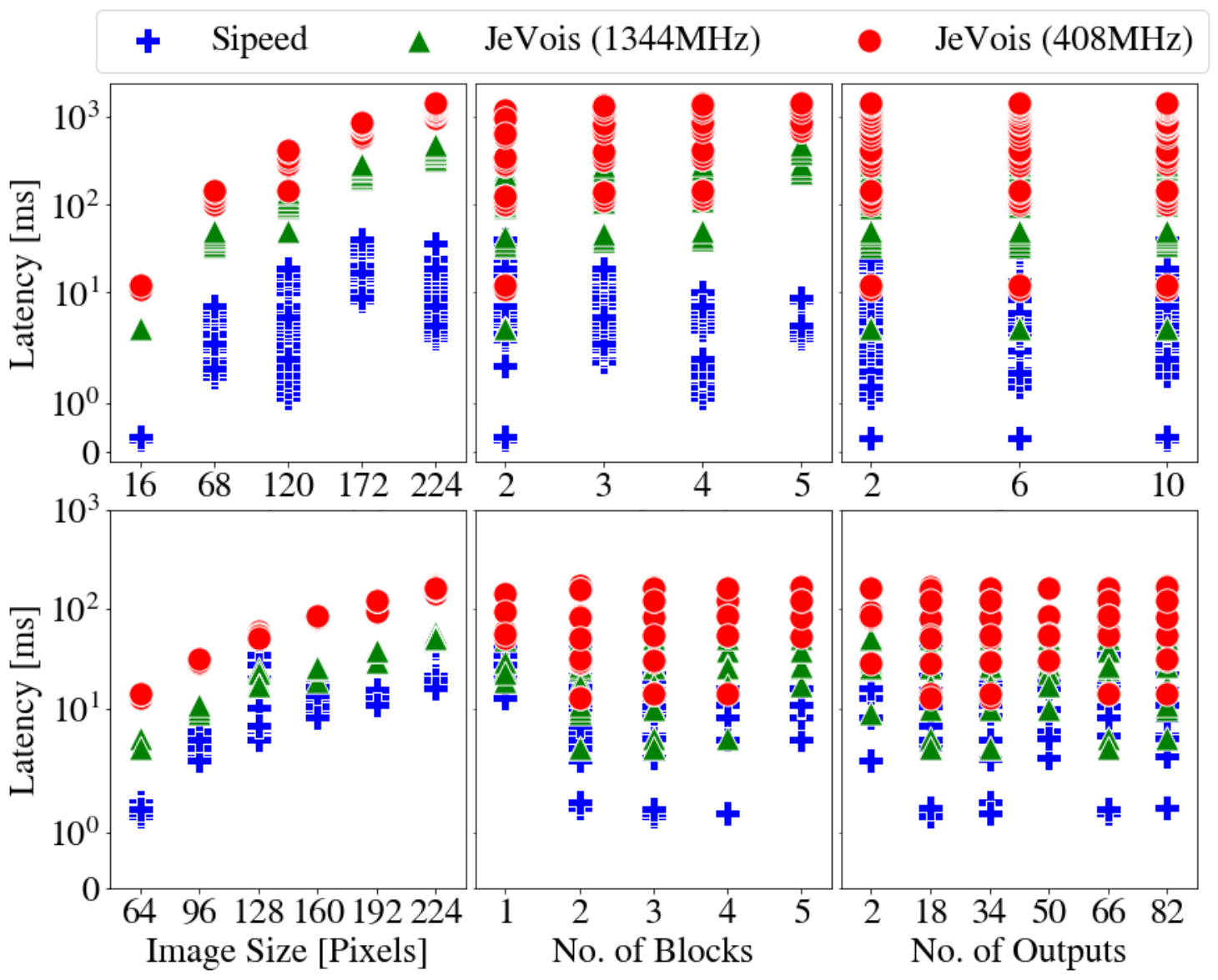}
	\caption{Latency comparison of two cameras, running Conv2D models(first row) and DepthwiseConv2D(second row).}
	\label{fig_conv2d_comp_r}
\end{figure}

Fig \ref{fig_conv2d_comp_r} compares the latency with respect to different hyper-parameters. Image size has direct correlation on latency for both cameras.
While the block size has inverse correlation for Sipeed camera. So its better to have large blocks for best performance.

\begin{figure}[h!]
	\centering
	\includegraphics[scale=0.26]{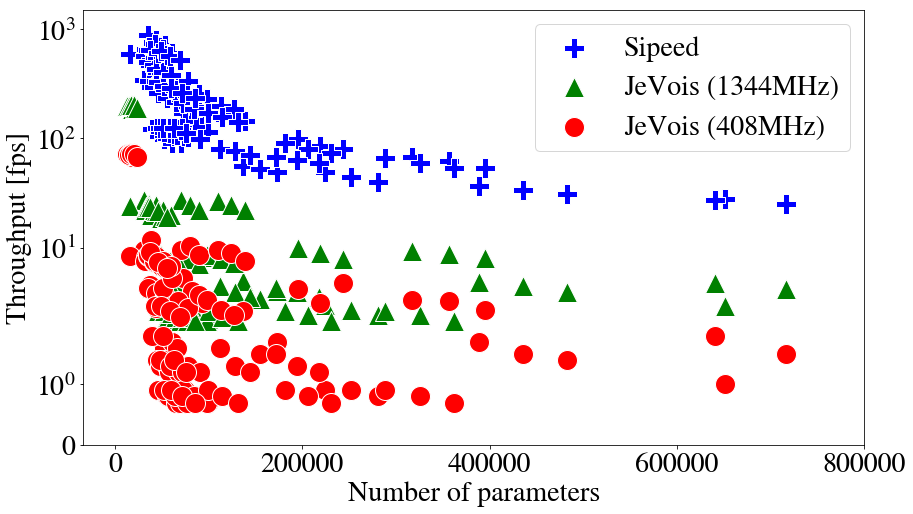}
	\caption{Throughput comparison of two cameras, running Conv2D models}
	\label{fig_conv2d_comp_f}
\end{figure}

\begin{figure}[h!]
	\centering
	\includegraphics[scale=0.26]{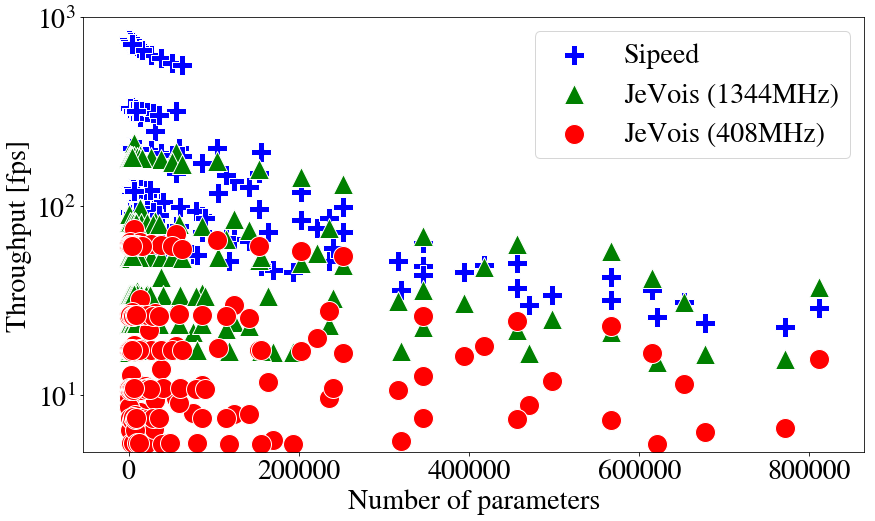}
	\caption{Throughput comparison of two cameras, running DepthwiseConv2D models}
	\label{fig_depth_comp_f}
\end{figure}

Fig \ref{fig_conv2d_comp_f} and fig \ref{fig_depth_comp_f} compares throughput values with respect to the number of parameters. Throughput is inversely related to the number of parameters for Sipeed camera. In case of JeVois camera, even though there exists an inverse relation some smaller models perform bad as compared to larger size models.

\subsection{Standard Architectures}
For second set of benchmark, we selected 2 popular embedded deep learning architecture MobilNets and YOLO for performance analysis.
MobileNet \cite{howard2017mobilenets} is a CNN architecture designed for mobile and embedded based vision applications where there is a lack of computing power. It provides high accuracy while the size of the network is relatively small. This architecture is based on depthwise separable convolutions, making it proper for building small models that can be matched to mobile and embedded vision applications. 
YOLO is a real-time object detection architecture designed for real-time processing. It uses a multi-scale training method to detect objects. It divides the image into a grid with bounding boxes, which are drawn around images. Predicted probabilities for each region are then calculated based on the weights that are associated with the probabilities\cite{redmon2017yolo9000}. 

Table \ref{table_standard_architectures} reports the latency and throughput values for both cameras when running MobileNet and YOLO. MobileNet deployed is of version 1 with 0.75x channel count, $224\times224$ input image size, and quantized weights while  YOLO consist of 20 classes detector with input image size $320\times240$ pixels and quantized weights.
As results show, Sipeed performs better than JeVois for both the models with respect to latency. While when comparing throughput Sipeed has higher throughput for MobileNet architecture but JeVois has higher throughput for YOLO architecture. This occurs because the last layer of YOLO is a convolution layer which increases the time taken to process the results from the final layer.

\begin{table}[h]
	\caption{Results of standard architectures experiment}
	\label{table_standard_architectures}
	\begin{center}
	\begin{tabular}{|l|l|c|c|}
		\hline
		Architecture & Camera & Latency [ms] & Throughput [fps] \\
		\hline \hline
		\multirow{3}{*}{MobileNet} & JeVois (408MHz)   & 397  & 3.3   \\ 
	      	                        & JeVois (1344MHz)  & 125  & 7.6   \\ 
		                            & Sipeed             & 38   & 26.2     \\ \hline
		\multirow{3}{*}{YOLO}       & JeVois (408MHz)   & 3275  & 113   \\ 
		                            & JeVois (1344MHz)  & 1320  & 140   \\ 
		                            & Sipeed             & 24   & 21     \\
		\hline
	\end{tabular}
	\end{center}
\end{table}

\subsection{Analysis}
Based on the performance analysis, we can conclude the following (i) Latency and throughput in both the cameras are directly related to the number of parameters. (ii) Sipeed performs better in terms of latency and throughput than JeVois, this because of the dedicated CNN accelerator but can only support limited architectures, (iii) JeVois on the contrary can execute a wide range of deep learning models with comparable performance.

\section{GRASP VERIFICATION}
\begin{figure}[t]
	\centering
	\includegraphics[width=0.8\linewidth]{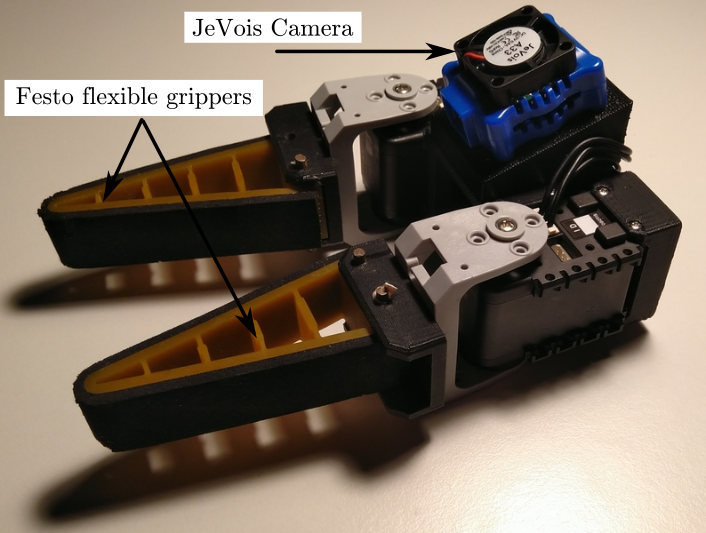}
	\caption{youBot gripper with mounted Jevois camera}
	\label{fig_gripper}
\end{figure}

Grasp verification task is formulated as an image classification problem where the camera perceives the gripper and it has to classify between 2 states "grasped" and "not grasped". For this purpose the JeVois camera is mounted on the youBot gripper (see Fig \ref{fig_gripper}), a custom dataset is collected, image classification is performed by using different CNN architectures, and the results are evaluated. The dataset contains more than 4000 images and sufficiently generalizes various grasping conditions. 

\subsection{System Design }
The limitations enforced by the hardware leads to a size limit on the deep learning models which in turn limits their learning capacity.
This limitation of the model is compensated in the integration of the machine vision camera. The different factors considered during the integration are (i) Field of view of the camera (ii) Force to be applied on the object being grasped and (iii) Lighting condition of the environment.
Field of view (FOV) of camera has a substantial impact on the complexity of the vision problem. A larger FOV causes information overload making it difficult to learn while a smaller FOV will cause loss of information required for the task. Since the major information of the grasp is in the gripper the camera is placed in such a way that the grippers are always visible during the grasp action. Force applied by the gripper has an impact on the bending of the gripper and is a good source of visual confirmation of a tight grasp therefore during the experiments the objects are grasped with the required force. Lighting conditions also affect the vision capabilities, therefore for our experiments normal industrial lighting conditions are assumed with illuminance ranging from 50 to 500 lux.

\subsection{Training and Evaluation}
The architectures selected for evaluation are MobileNet, single block ResNet, and a custom CNN architecture based on the insights from the benchmarking experiment. Tensorflow was used to model the architectures and the training was performed offline on an Intel CPU.

\subsubsection{MobileNet}
Fine-tuned MobileNet model is created by modifying a MobileNet model with 0.5x channel count that is pre-trained on ImageNet dataset. The last 6 layers of the model is removed, then the last 12 layers are trained on the collected dataset. The final model contains 830,562 parameters with 401,922 trainable parameters. 

\subsubsection{Single Block ResNet}
ResNet architecture follows a framework called deep residual learning that is designed to solve the problem of degradation. To turn a plain network into a residual network, shortcut connections are inserted between groups of layers \cite{he2016deep}.
For this experiment, the model is reduced to a single block ResNet architecture. Final model contains total 87,170 parameters with 85,762 trainable parameters. 

\subsubsection{Custom CNN Models}
Here two CNN models that were introduced in benchmarking are trained. These models are constructed from simple CNN blocks, they are small when compared with MobileNet or ResNet architectures. The model with Conv2D layers contains 66,194 parameters, and model with DepthwiseConv2D layers contains 248 parameters, all trainable.

\begin{figure}[b]
  \centering
  \begin{subfigure}[b]{0.49\linewidth}
    \includegraphics[width=\linewidth]{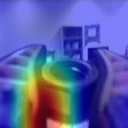}
    \caption{Grasped}
  \end{subfigure}
  \begin{subfigure}[b]{0.49\linewidth}
    \includegraphics[width=\linewidth]{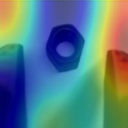}
    \caption{Not Grasped}
  \end{subfigure}
  \caption{Saliency Map of MobileNet trained model, darker pixels contribute more towards prediction.}
  \label{fig:saliency_map}
\end{figure}

\begin{table}[h!]
	\caption{Summary of model training}
	\label{table_train_all}
	\begin{center}
	\begin{tabular}{|l|l|c|c|c|c|c|}
		\hline
		Architecture & Parameter & Train & Validate \\
		\hline \hline
		\multirow{2}{*}{MobileNet}      & Accuracy  & 0.98     & 0.97  \\ 
		                                 & Loss      & 0.03     & 0.12  \\ \hline
		\multirow{2}{*}{ResNet}          & Accuracy  & 0.94     & 0.97  \\ 
		                                 & Loss      & 0.13     & 0.08  \\ \hline
		\multirow{2}{*}{Conv2D}          & Accuracy  & 0.98     & 1.00  \\ 
		                                 & Loss      & 0.03     & 0.01  \\ \hline
		\multirow{2}{*}{DepthwiseConv2D} & Accuracy  & 0.83     & 0.95  \\ 
	                                     & Loss      & 0.39     & 0.21  \\ 				                              
		\hline
	\end{tabular}
	\end{center}
\end{table}

\begin{table}[h!]
	\caption{Summary of models performance}
	\label{table_thr_lat_youBot}
	\begin{center}
	\begin{tabular}{|l|c|c|c|}
		\hline
		Architecture & Latency [ms] & Throughput [fps] & Parameters \\
        \hline \hline
        MobileNet & 75 & 12 & 830,562 \\ \hline
        ResNet &  140 & 6.9 & 87,170 \\  \hline
        Conv2D &  160 & 5.8 & 66,194 \\  \hline
        DepthwiseConv2D &  17 &  43 & 248 \\
		\hline
	\end{tabular}
	\end{center}
\end{table}
\begin{figure*}[t]
	\centering
	\includegraphics[width=\linewidth]{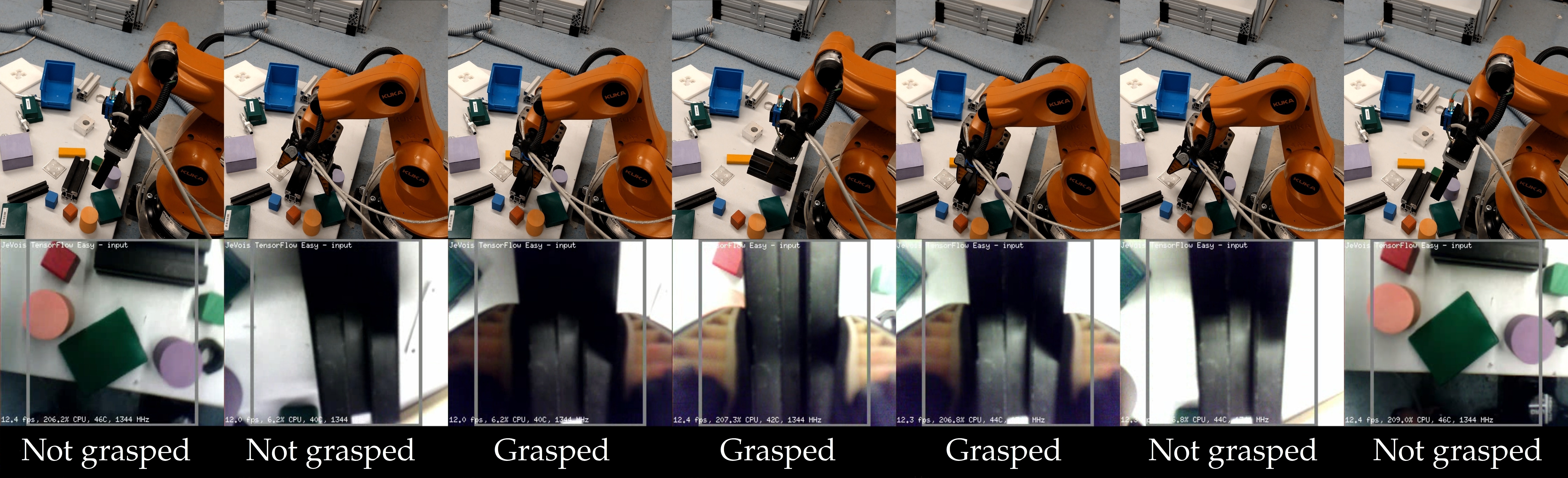}
	\caption{youBot performing pick action, along with the inferred grasp verification status. First row: from observer's view, second row: from camera's view.}
	\label{fig_youBot_steps}
\end{figure*}

MobileNet achieves 98\% accuracy in training and 97\% accuracy in validation. Even with larger number of parameters it has the lowest latency and throughput as compared to ResNet or Conv2D. 
On analysis of the saliency map as show in fig \ref{fig:saliency_map}, we observe that the gripper curvature and gripper tip are the most important features contributing towards the decision. 


\subsection{Evaluation of Integrated System}
To investigate the performance of the trained model on the overall integrated setup, we selected several objects with varying shape, weight and texture. For each object, three distinct grasp variations were tested (i) grasping with different forces (ii) grasping with different backgrounds and (iii) grasping in varying lighting conditions. In each of these variations, the robot picks each object from different picking positions, lifts it and then verifies if the grasp was successful. Fig \ref{fig_youBot_steps} provides an overview of a single run of the integrated system experiment. Based on the training accuracy we selected MobileNet architecture to run inference on the JeVois camera. 
The integrated system was able to verify the grasp for all the objects with 97\% per frame accuracy and 100\% per run accuracy.

\section{CONCLUSIONS AND FUTURE WORK}

This paper provides a vision based deep learning solution for grasp verification using machine vision cameras. We comprehensively benchmarked deep learning inference capable machine vision cameras. Based on the benchmarking results, the JeVois camera was integrated to the KUKA youBot gripper for grasp verification. The dataset collected from the integrated setup was then used to evaluate the performance of four deep learning architectures. Finally, the trained MobileNet-based grasp detection was deployed and evaluated with different test objects, in which it achieves 97\% per frame accuracy and 100\% per run accuracy.
The dataset, generated models and other benchmarking results are openly available \footnote{https://github.com/amirhpd/grasp\_verification}.
One future work is to increase the capability of the network to detect other semantic information, such as grasp quality and slippage by redefining the model architecture considering the benchmarking results of the cameras.

\addtolength{\textheight}{-12cm}   


\section*{ACKNOWLEDGMENT}
\small The authors gratefully acknowledge  the  on-going  support  of  the  Bonn-Aachen International Center for Information Technology. 
\bibliographystyle{./IEEEtran} 
\bibliography{bibliography.bib}

\end{document}